# Temporal Reasoning About Uncertain Worlds *

Steve Hanks

Department of Computer Science, Yale University

P.O. Box 2158 Yale Station, New Haven CT 06520

June 6, 1987

## Abstract

We present a program that manages a database of temporally scoped beliefs. The basic functionality of the system includes maintaining a network of constraints among time points, supporting a variety of fetches, mediating the application of causal rules, monitoring intervals of time for the addition of new facts, and managing data dependencies that keep the database consistent. At this level the system operates independent of any measure of belief or belief calculus. We provide an example of how an application program might use this functionality to implement a belief calculus.

## 1 Introduction

How to reason with time—what state the world is in, what states it might take on in the future, and how the past and present affect what is to be—is something that most of our theories must take into account, and that most of our programs must do somehow. Furthermore, in all but the most simplistic and constrained worlds one must also deal with the fact that one's knowledge of the world is rarely certain. One typically doesn't know exactly what state the world is in, or exactly the effect one's actions will have when they are actually performed. And even if one *is* reasonably sure about these factors, there are often other actors or autonomous forces in the world that must be taken into account. But even given the ubiquity of temporal reasoning in solving planning, scheduling, and projection problems, and the inherent uncertainty of these domains, we know of no efforts to incorporate an

*This reasearch was supported by DARPA/BRL grant DAAA15-87-K-0001. Thanks to Drew McDermott, Jim Firby, and Lucian Hughes, who were their usual helpful selves.

explicit model of uncertainty into a temporal representation or into a program that reasons with a rich model of time.

We will discuss in this paper a research effort directed toward this goal: a program that manages a temporal database in which facts are not certainly true and in which one's knowledge is both incomplete and dynamic. We are building this temporal database manager to support a plan projection/evaluation program. The projection program, which evaluates the effects of performing actions in an uncertain world, is itself part of a planning system that we are developing at Yale. In the first part of the paper we will motivate the research by looking at the bigger picture—the problems involved with temporal reasoning, plan projection and planning in an uncertain world. The rest of the paper will be spent describing the philosophy behind and the implementation of the temporal belief manager (TBM) itself.

## 2 Planning and Plan Projection

The problem of plan projection might be stated as that of reasoning about the expected ramifications of taking some action or series of actions, given what is believed about the world, what actions one has previously committed to, and what are believed to be the immediate effects of the actions. What we want from the projector is some small set of scenarios—ways the world might turn out if the contemplated actions were actually taken. The scenarios should be in some sense "significant," or "interesting." There will typically be a great number of ways the world might turn out, and we don't care about most of them. Of course a reasonable definition of significance is very difficult, and we will not pursue one here. It is clear, however, that judging the significance of a scenario involves a tradeoff between the likelihood that the scenario will be realized and the impact that such an outcome would have on the planner's goals. Such is the province of decision analysis, though it is unlikely that a projection program would be able to undertake a rigorous analysis in the manner described by Raiffa [12]. In a reasonably complex planning domain we will not be able to describe exhaustively the ways the world might turn out in response to our actions, and a formal characterization of the utility function will tend to be difficult as well. For the moment we are worrying about how to make the risk and impact information available to the planner quickly; what it then does with the information is the topic of future papers.

Automated planning systems have not dealt with the projection problem in any generality. Instead they have either recognized it but not attempted a solution (Wilensky [16], Hammond, [8]), or introduced into the planning paradigm assumptions that made it much easier (Fikes [6], Sacerdoti [13], and Firby [7]). The focus of our attention will be this last group—the series of reductionist planners including STRIPS, NOAH, NONLIN and FORBIN.

The approach taken by reductionist planners can be characterized as follows: given a goal, they retrieve from some plan library a plan that will achieve the goal. This plan may contain primitive actions (that may be performed in the world) and subgoals, which in turn must be achieved. The plan is entered into a data structure called the *task network*, which records the planner's goals, the plans and actions it has chosen in order to achieve those goals, the subgoals spawned to enable the subplans, and so on. (See, for example, Charniak and McDermott, [3, chapter 9].) The planner then goes on to plan to achieve the new subgoals and to order the actions so as to eliminate unwanted interactions. The task network records the planner's commitments to perform actions as well as what it knows about the planning world, thus contains at least part of the information needed to do plan projection.

And it turns out that due to certain very stringent certainty assumptions made (implicitly) by these planners, the task network is *all* that's necessary to do plan projection, and in fact maintenance of the task network *is identical to* projection. This state of affairs conflates several concepts that need to be considered separately when the certainty assumptions are relaxed.

First we should list those assumptions. The first is that the state of the world is known exactly and completely at the beginning of the planning task. The second is that the effects of all actions executed at the behest of the planner are known exactly. And third is that the planner is the only thing that makes changes to the world; that is, there are no other planning agents or autonomous forces at work in the planning world. (Or at least if there are, their actions and the effects of those actions are known with certainty.)

These assumptions lead to an interesting state of affairs. Most notably, there is no particular problem about maintaining a view of the world: the task network always contains a complete and accurate record of every fact of interest to the planner. The distinction between planning time and execution time is similarly obscured: there is no need to wait and see if a particular plan will work when it is executed, in that all the information needed to make that judgement is available when the plan is built. And therefore the problem of plan projection (in the sense of reasoning about the effects of the planner's actions on the "real" world) is no different from building the plan in the first place. Both depend only on the task network (and not, for example, on observations gathered from the external world), and both can be done at plan-construction time.

Of course when the certainty assumptions are relaxed these distinctions become apparent, and projection becomes an interesting problem in its own right. At Yale we are reconsidering the paradigm of reductionist planning in a domain that admits uncertainty in a limited fashion. We are considering an autonomous mobile supply robot that travels along known roads and plans to achieve goals like "have 20 barrels of gasoline at the fuel depot by 9:00 tomorrow morning." We relax all three of the uncertainty assumptions above. At any point in the planning process the planner will have only limited information about the state of the world: it can't be certain, for example, that the fuel drums will be in the loading dock where it expects to find them. The effects of actions may not be known with certainty: the amount of time it takes to traverse a particular road is a random variable whose distribution depends on the speed of the truck and other factors like the time of day and the weather. Autonomous processes, like the weather, operate in the planner's world and affect the outcome of the planner's actions. The behavior of autonomous processes can be forecasted with varying degrees of precision and at varying costs.

In such a world there is a distinction between commiting to perform a certain task and actually performing the task, thus a distinction between planning and execution and between the task network and the planner's view of the world. In our planning architecture the planner maintains the former and the projector maintains the latter. (Of course the two aren't completely separate—the planner will typically need to know the expected state of the world in order to choose appropriate plans.)

As we mentioned above, one of the main responsibilities of the the projector is to maintain a view of the planning world. Because of the uncertain nature of the world the projector keeps a record of what the planner *believes* rather than what is true; because of the dynamic nature of the world these beliefs tend to change over time, both due to the addition of new information and because of the tendency of information to "decay" over time. This paper discusses the machinery necessary to support reasoning of this sort. First we discuss existing systems that do temporal reasoning and belief maintenance, in order to point out the functional needs and justify the application-level functionality of our system. Then we describe the data types and procedures that implement that functionality, and provide an example.

## 3 Temporal Reasoning and Uncertainty

Three main problems concern programs that support temporal reasoning (for example the work of Dean [4], Allen and Kautz [1], and Williams [17]): one is to keep track of the temporal relations that hold between what is known, the second is to respond to queries about what is known based on these relations, and the third is to support inference—to add to what is known based on the old facts and their temporal proximity. Most systems have concerned themselves with the first problem—maintaining a consistent set of temporal constraints and processing additions, deletions and queries efficiently. We will speak mostly about Tom Dean's work here because he treats the problem of fact retrieval and inference more completely than the others; the discussion of constraint maintenance applies to the other systems as well.

All of these temporal reasoning systems tolerate uncertainty to some limited degree, in that relationships among the temporal objects (time points or intervals) can be under-specified. Allen's system keeps track of relationships between time intervals, like *equals*, *before*, *after*, *overlaps*, *etc*. For any two intervals the system maintains the *set* of relationships that can hold between them, given the current set of constraints. Thus uncertainty is represented in the implicit disjunction. Similarly, Dean's system records the distance between two time points as a range of numbers, allowing the possibility that one is uncertain about the order in which two points occur temporally.

So the uncertainty in Dean's temporal reasoning system is that inherent in the partial ordering of time points imposed by the constraints—uncertainty is permitted about *when* things are believed, but not about *the extent to which* things are believed. Within that framework he provides several methods of resolving or reasoning about that uncertainty. Two such mechanisms are "fact persistence" and "default constraint queries."

Dean's "rule of persistence" is a simplified implementation of McDermott's notion of fact persistence in [10]. The basic idea is that when one asserts that a fact begins to be true at some point in time, one typically doesn't know when it is going to *stop* being true. In the persistence model a fact tends to remain true (forever, in Dean's implementation) unless and until the fact is contradicted by a contradictory fact. So the rule of persistence is a way of resolving the uncertainty in the partial order of time points by letting the endpoints of facts lie as temporally late as consistently possible. Below we present an implementation of a more sophisticated model of persistence.

Default constraint queries allow the program to reason hypothetically about the possibility that the partial order will resolve itself in a certain way. The application program asks the database manager "is it consistent to believe that point $p_1$ occurs later than point $p_2$?" and if this query is consistent with the current partial order the database manager returns a "yes" answer and sets up data dependencies ensuring that the application program will be notified if this assumption is violated. This provides the application program with a way of reducing uncertainty in the system through making assumptions. Our system again provides a more general notion of associating beliefs with statements about the constraint network.

Dean also speaks about dealing with uncer-tainty by maintaining and reasoning about "partial world descriptions," which are sets of alternative planning options, or courses of events, associated with the current partial order. This is close in spirit, but different in implementation from our model of reasoning in hypothetical worlds that we will discuss in the last section.

## 4 Belief Maintenance

Now that we've discussed programs that reason about time we should turn our attention to programs that reason about belief. There is less to talk about here just because the uncertainty people have not had to deal with temporal issues in the same way the temporal people have had to deal with uncertainty. Uncertainty seems to creep in to temporal reasoning no matter how hard one tries to assume it away, while those that deal in beliefs and uncertainty can assume time away and still get some interesting work done.

The uncertainty-related issues we have to deal with are the usual ones: what is a a reasonable way to express uncertainty, belief, ignorance? How does one combine old beliefs, which may be interdependent or inconsistent, into new beliefs, and do so quickly.

Among the systems for the domain-independent maintenance of a network of beliefs, for example Pearl [11], Tong *et. al.* [15] and Falkenhainer [5], our work is most similar to the last. Our system, like his, tries to make available to an application program the machinery necessary to support probabilistic inference, without committing to a particular representation or combination paradigm. In fact, our work might be superficially characterized as an extension of Dean's Time-Map Manager to include an explicit representation of uncertainty, or an extension of Falkenhainer's Belief-Maintenance System to include an explicit representation of time.

## 5 System Overview

Now we can begin an overview of the TBM program, which we will do by describing the data types it defines and the operations it performs on them. We will start with the temporal constraint network, then work our way up to computing beliefs and performing inference.

### 5.1 Temporal objects

The basic temporal object is the *time-point*, which denotes an instant of time. The distance between two points is stored as a pair of numbers (which may be infinite) representing the belief that the actual distance between the two points lies somewhere in the interval defined by the range. An application program constrains temporal distance by posting *constraints*, which involve a pair of points and a pair of numbers indicating the distance. *Time-intervals* denote a continuous stretch of time, and are built out of time points.

A novel idea is that of a temporal *overlay*. An overlay is a partition of time into packets of an application-supplied "grain size," and contains points, intervals, beliefs, *etc*. The application program defines these overlays, which are supposed to reflect the structure of its problem-solving task: the idea is that points within an overlay are typically interrelated (and therefore the application program will probably be interested in the distance between them), but that points in different overlays are probably not (thus a distance estimate may be less accurate and harder to come by). By specifying a grain size the application tells the TBM that temporal durations shorter than this duration are insignificant. The TBM guarantees to return inter-point distance estimates to within the accuracy of the grain size, but no better. Computing these distance estimates, and adding new points and intervals to the overlays, are both extremely fast.

The other advantage to overlays is that they limit the number of objects that need to be searched during fetches. The application may ask the system to fetch all the beliefs of a certain type, for example, and in doing so must supply a set of overlays in which the fetch is to take place. Of course some beliefs may be missed if the application doesn't supply the right ones, but it can control the tradeoff between how long the fetch takes and how likely the system is to find all instances. It is also able to direct the system toward those overlays that are most likely to contain the right beliefs.

Note that overlays are somewhat different from Allen's reference intervals [1]. Overlays need not be disjoint, either temporally or in the points they contain. In fact we expect it would be common for an application program to view a certain stretch of time through more than one overlay—one with a large grain size for higher level plan-ning and one with a smaller grain size for detailed scheduling. Time points introduced in the course of low-level plan expansions would thus never appear in the coarser overlay. Overlays seem to offer the same efficiency advantages as reference intervals, but are a good deal more flexible.

From here on, when we speak of adding a point or a belief to the system, or fetching beliefs of a certain type, we will implicitly mean "with respect to a set of overlays."

## 5.2 Beliefs, monitors, and inference rules

The data type *belief-instance* represents belief in a *type* over a *duration*, and with a *strength*. The type is an S-expression, the duration is a time point or interval as defined above, and the strength is a quantity representing strength of belief, which is defined and manipulated by the application program. Belief instances have some other information that allows them to be updated as the system changes. The first is a list of objects (typically other beliefs or inference rules) that the belief depends on—that is, the objects were used to calculate the belief's strength. Next is the belief's *signal-function*—a function that is called when any of the objects it depends on changes in some way. Thus the belief can adjust itself to changes in the network. It is generally the application program's responsibility to build these belief instances (including the signal functions); the TBM in turn makes sure that the belief instance is placed correctly in the temporal network, and makes sure that it is signalled when any relevant conditions change.

The system supports one more kind of belief—beliefs about point-to-point distances. the application can ask the TBM to assess the belief that the distance between two points is in a certain range. The TBM then computes its best current estimate of the distance, and passes it to an (application-supplied) function that computes a belief strength. The system then posts this belief and sets up the data dependencies ensuring that the belief's signal function will be called if the distance assumption is ever violated.

*Monitors* represent "absence of belief" in some sense. One may base belief in a proposition on lack of belief in some other proposition. For example, I may believe that my tennis match this afternoon will indeed be played because I believe that it will be sunny at 5PM. This belief may in turn be based on the fact that it was sunny this morning, and also that there will be no rainstorms between now and 5PM. One way to represent this state of affairs is by installing a monitor, which looks for beliefs in facts of type "rainstorm," over the interval between now and 5PM. The monitor has a function that will be called if any such beliefs are added, and should then call the signal function of my belief in "sunny at 5PM" (which may in turn call the signal function of my belief in "tennis match will be played." Monitors correspond to the idea of "anti-protection" that Dean speaks about in [4]. The implementation is problematic in his time-map manager, but fits nicely and runs efficiently in our overlay-based implementation.

*Causal-rule*s are the way the application program gets the TBM to do inference. Rules are temporally scoped (so they apply only to specific periods of time) and are added to overlays (so as to limit the amount of work that needs to be done to monitor their firing). The significant slots in a causal rule object are the *precondition* patterns, the *trigger* pattern, and the *consequent* pattern. The basic idea is that if a rule spans a certain interval of time in an overlay and a trigger event occurs at some point in time when all the precondition patterns are true, that should generate a belief in the consequent pattern at the time the trigger occurs.

Patterns are S-expressions, and may contain variables under the restriction that any variables in the precondition patterns must also appear in the trigger pattern. Thus once an occurence of the trigger pattern has been identified the system can determine whether a particular rule should fire just by searching for ground instances of the precondition patterns.

In addition to the various patterns, the application must also supply a function that decides on a belief strength for the consequent, given beliefs in the preconditions and in the trigger (and possibly other information as well).

Much of the work the system does in maintaining these rules involves trying to ensure that the rules fire whenever necessary, but never unnecessarily. The algorithm goes something like this: when an application wants to know the belief in a pattern that is the consequent of a rule (or more precisely, can be unified with the consequent of a rule) the system tries to produce all instances of the consequent within a period of time preceding the request by the *consequent-duration*—a length of time supplied by the application as part of the rule. It assumes that any firings prior to that are of no interest. First it looks for all instances of the trigger pattern (suitably instantiated) that occur in that interval. Then for each such instance it assesses its belief in all the precondition patterns at the point of the trigger's occurence. It then calls the rule's generator function (mentioned above) that builds a belief instance for the consequent.

In addition to creating the new instance the system sets up dependencies to make sure that the rule is subsequently fired if its preconditions are met. The assumption is that since the application asked about the consequent pattern in an interval it will also be interested in subsequent rule firings in that interval. For each rule the system sets up a monitor that will look for new assertions of the trigger pattern, and go through the same process of verifying preconditions if one is discovered. Further, if it originally found a trigger belief but did *not* add an instance of the consequent (presumably because the preconditions were not believed strongly enough), it sets up dependencies that ensure that it will try again to add one if belief in the preconditions change. Thus once the application expresses an interest in a pattern that is the consequent of a rule, it can be assured that it will be informed whenever the rule can be fired.

### 5.3 Summary of functionality

We've already discussed most of the system's basic functionality in introducing the data types. Another way of looking at the system is as a base-level functionality on which to build a belief calculus. Here we summarize the base-level functionality and make clear what the application program needs to provide to these routines in order to compute beliefs. This is what the base-level routines do:

MANIPULATE THE CONSTRAINT NETWORK: you can define time points and time intervals, and add them to overlays. You can impose distance constraints on time points.

FETCHING: you can perform a variety of fetches in overlays, but the most important one is to fetch all belief instances of a particular type that occur over a particular interval of time.

BELIEFS: you can make a flat assertion of belief—"I believe this fact with this strength at this point in time" or you can ask the system to assess your belief in a fact at a point in time (but this requires some functions from the belief calculus, as detailed below). You can get an assessment of the extent to which you believe that a particular temporal ordering condition holds.

RULES: you can define causal rules, as defined above, and expect that when you then fetch on the consequent rule type the fetch will cause the rule to fire, and furthermore it will fire in the future if the preconditions are met in the interval of interest.

In turn, the base-level routines depend on the application for the following information:

A DEFINITION FOR THE BELIEF-STRENGTH DATA TYPE. This might be a number that takes the value 1, 0, or -1, or perhaps a real number representing a probability, or perhaps a pair of numbers representing a "Shafer interval" [14]. The base-level routines don't care.

AN ASSESSOR FUNCTION FOR FACT TYPES. The application must define for each fact type an "assessor function," which is called whenever the application asks the system to assess its belief in that fact type. The assessor function takes as arguments the fact type itself and a set of constraints, and returns an appropriate belief instance. The function is of course free to call the base-level routines, in particular to fetch for temporally prior beliefs. The assessor function must also build a signal function for this belief—the signal function will be called whenever a change in the belief network means the belief should be reconsidered.

CAUSAL RULE DEFINITIONS. For each such rule the application must supply precondition, trigger, and consequent patterns, as well as the time interval over which the rule is to be active. Additionally the application must supply a "generator" function, which computes the strength of belief in the consequent, given beliefs in the preconditions and belief in the trigger.

Next we'll look at how we can use these routines to build a simple belief calculus.

## 6 Example: A Simple Belief Calculus

As an example we will present a problem we have used in the past (*e.g.* in [9]) to explore the way certain formal systems handle temporal projection problems. The example involves three causal rules and three events. (Events in our system are just belief instances asserting that the event occured.) Informally put, they are:

1. Rule 1: If a person is born then that person begins being alive.

2. Rule 2: If a gun is loaded then that gun begins being loaded.

3. Rule 3: If a person is shot with a loaded gun then that person stops being alive.

4. Event 1: FRED is born.

5. Event 2: Twenty years later, THE-MURDER-WEAPON (a gun) is loaded.

6. Event 3: One hour later, FRED is shot with THE-MURDER-WEAPON.

Given these rules and events, the question is what do we believe about FRED's being alive at some point in time after the shot. We need to make several decisions in representing this problem: what is our measure of belief? How do we compute the causal relationship between a "born" event and being alive, between a "load" event and being loaded, and between a "shoot" event and being alive? How do we assess our belief about someone's being alive on the basis of prior beliefs, and similarly our belief that something is loaded based on prior beliefs? We will provide some answers below, such that our program can actually produce meaningful results. We stress however, that these decisions are going to be pretty arbitrary and simplistic—the point is to show in a brief example how the program works.

First the belief strengths: we will just make an arbitrary decision and define a belief strength in the same way the Falkenhainer does (in [5])—as a "Shafer interval" of the form $[s(P), s(\neg P)]$, where $P$ is the fact type of the belief instance, and s is a function mapping a fact type into a real number in the interval [0,1] that represents the extent to which evidence supports belief in its argument. Thus the first number represents the extent to which evidence confirms the fact and the second represents the extent to which evidence disconfirms it. Thus we represent negation implicitly: evidence confirming $P$ is evidence disconfirming $\neg P$, and *vice versa*.

Next we'll tackle the persistence of "alive" facts. We're going to try one simple model of persistence here: let's say that knowing that someone was born counts as a certain amount of evidence that the person is alive, as long as that person is younger than 70 years. If he's older it's no evidence. But we can get evidence of "aliveness" from other sources as well: seeing the person alive, seeing the person dead, reading his obituary in the *National Enquirer*, etc. We will assume that pieces of evidence of this sort, which we will call *observations*, are entered directly into the database as belief instances, thus have belief strengths already associated with them. We then have to decide how fast the evidence supplied by these sources decays over time, and also how to combine several observations into a single belief strength pair.

There's a significant asymmetry between being alive and being dead. One tends to stay dead, but not to stay alive. Thus evidence of one's being dead tends to retain its value over time, but not so for being alive. We can thus assume that the *second* number of our belief strength will stay the same over time. Not so for the first: if I observed somebody alive a week ago that's pretty good evidence that he's still alive, but if it was five years ago it's a different matter. We'll just assume that the evidential value associated with being alive decays linearly from the time of observation such that six months later such an observation provides no information. Finally, to combine these sources of evidence (his "initial aliveness" and any additional observations we may find in the database), we will make the usual independence assumption and use Dempster's rule.

That's all the information we need to write the assessor function for fact type "alive". Recall that the assessor function takes in the fact type and a set of constraints (representing the time of the query), and returns a belief instance. We're most interested in computing the belief strength of that instance, and here's how we do it: we first need to search back from the time of the query (but no further back than the evidence lifetime of six months) to find instances mentioning the fact type (*(ALIVE FRED)*, or something like that). We can find two sorts of belief instances: observations, and assessments. The first represents actual sources of evidence, and the second represents computed measures of our belief. We discard everything prior to the latest assessment, leaving us with (at most) one assessment and various observations that have happened since. We then "discount them to the present" by decaying the evidence supporting "aliveness" according to the function above. We also need to add in the evidence that "under 70 indicates being alive," so we have to search backward in time looking for the person's birth and noting whether it occured more recently than 70 years ago. (This is a horrible state of affairs, but necessary because we're not representing the concept of age.) Finally we apply Dempster's rule, which produces a belief strength. We also have to note in the

120

belief instance itself those beliefs that our calculation depended on, so we will be notified if we should reconsider the calculation.

The persistence of the "loaded" fact is easier, due to a different persistence assumption we will make. The assumption is that *any* evidence of a gun being loaded or unloaded tends to decay over time, gravitating toward a strength of [0,0]. We will assume that the decay is linear over the period of one month. Thus the assessor function for fact type "loaded" is as follows: search back over the past month to find beliefs in "loaded" (it doesn't matter in this case whether they are observations or assessments). Take the latest one, if any. (If not, the strength is [0,0].) Decay the belief strength of that instance from the time of the observation to the time of query. This becomes the new belief strength. In addition to the new belief instance we also need to post a monitor to the database, asking to be notified if any new "loaded" facts are added between the latest observation and the query, so we can recompute the belief strength.

After all the difficulty building the assessor functions, building the causal rules is surprisingly easy. We will describe only the third one: "getting shot with a loaded gun is evidence for not being alive." The main thing we have to is write a function that computes a belief strength for being alive based on beliefs that the gun was loaded and that the shot really occured.

Both of these quantities are passed as arguments to the function. Assume that the strength of belief in "loaded" is $[l_1, l_2]$, and the strength of belief in "shoot" is $[s_1, s_2]$. We interpret $l_1$ as the probability that the gun was loaded, and $s_2$ as the probability that the shot took place, thus return the quantity $[0, l_1 s_1]$ as the evidence that the victim is alive *as the result of the shot*. This piece of evidence has to be combined with the evidence that the victim was alive *at the time of the shot*. The latter evidence we get through the normal assessment process (*i.e.* we ask the fetch routines), and we combine the two using Dempster's rule.

Our program runs two variants of the above example. In the first case, all the event occurences are added (with certainty) then the system is asked to assess the likelihood that FRED is alive. It reports evidence that he is alive based on the fact that he was born some twenty years before, and evidence that he is dead based on the occurence of the shot. In the second case we omit the "load" event and the system reports no evidence that he is dead (because it has no evidence of the fatal shot, because it has no evidence that the gun was loaded). We can then add the "load" event, at which point all the appropriate monitors fire, belief strengths are altered, and the system decides that he might have died after all.

## 7 Extensions

We should first mention where future work needs to be concentrated. As far as program implementation, what we need next is a way of reasoning about "hypothetical worlds." In contrast to Dean's system, which supports simultaneous reasoning about partial world descriptions, we are experimenting with a system where the world "splits" on the basis of a set of assumptions (these being constraints placed on the values that belief strengths can take on), and the application then explores these hypothetical worlds explicitly and separately. Efficiency is the main issue here, in that the application must be able to create and destroy worlds fairly quickly (though not indiscriminately).

Much work needs to be done on the theoretical basis for building a belief calculus. as we did in the example. We made many arbitrary assumptions in our example, which must eventually be put on a firmer theoretical footing. The most important one has to do with combining evidence from disparate sources—for example the evidence that the gun was fired with the evidence that the gun was loaded with the evidence that the victim was alive at the time of the shot.

Add to that a problem that we ignored altogether in the example: the possibility that point orderings might lead to ambiguity. In our example the events were totally ordered, but if we that condition does not hold we have to add to the above calculation the evidence that the loading event occured before the shot. Currently the assessment of beliefs based on information from the constraint network is done in an *ad hoc* manner.

## 8 Conclusion

What we hope to have accomplished in this paper is to motivate the problem of temporal reasoning under uncertainty—to show how it comes up in the planning process and why it can't be handled either by systems that do temporal reasoning or by those that do belief maintenance.

We feel that our program is a first step in solving this problem; while many technical issues remain unsolved, we have demonstrated that the functionality our system offers is adequate to solve nontrivial problems in the domain.